# Two-Layer Mixture Network Ensemble for Apparel Attributes Classification

Tianqi Han, Zhihui Fu, and Hongyu Li*

AI Lab, ZhongAn Information Technology Service Co., Ltd.
Shanghai, China
{hantianqi,fuzhihui,lihongyu}@zhongan.io

**Abstract.** Recognizing apparel attributes has recently drawn great interest in the computer vision community. Methods based on various deep neural networks have been proposed for image classification, which could be applied to apparel attributes recognition. An interesting problem raised is how to ensemble these methods to further improve the accuracy. In this paper, we propose a two-layer mixture framework for ensemble different networks. In the first layer of this framework, two types of ensemble learning methods, bagging and boosting, are separately applied. Different from traditional methods, our bagging process makes use of the whole training set, not random subsets, to train each model in the ensemble, where several differentiated deep networks are used to promote model variance. To avoid the bias of small-scale samples, the second layer only adopts bagging to mix the results obtained with bagging and boosting in the first layer. Experimental results demonstrate that the proposed mixture framework outperforms any individual network model or either independent ensemble method in apparel attributes classification.

## 1 Introduction

Automatically recognizing apparel attributes has recently drawn great interest in the computer vision community. The recognized apparel attributes can be used in various applications, for example, automatic product tagging, clothes searching, clothing style recognition and clothes matching strategy learning. However, the annotation of the clothes requires special fashion domain knowledge and careful data cleaning, so that the available datasets are quite limited, thus the trained model is leading to overfitting with a high probability. In the meanwhile, clothes, with high variation including deformation and occlusion, require that the recognition models have better capabilities of describing clothes features.

Most previous clothing recognition methods [3, 4] are based on the hand-designed features which are hardly optimal for the customized classification tasks. For deep learning methods, DeepFashion [2] was proposed recently to handle clothing deformations and occlusions via jointly predicting the landmark locations and clothing at-tributes which achieves robust performance on the clothing recognition problem. Methods based on various deep neural networks for image classification can also be applied to apparel attributes recognition. An interesting problem raised is how to ensemble these methods to further improve the accuracy of apparel attributes classification.

Ensemble can be regarded as a classification task based on the outputs from all the base predictors, which will mine the correlation among all the dimensions of one base predictor output and the inter-predictors' relationship. Generally speaking, ensemble methods can smooth the outputs of multiple classifiers and reduce the variances. Tree boosting techniques have demonstrated good performance in classification tasks. Recently, gradient tree boosting classifiers [1, 10] are proposed as a functional gradient descent problem which has given state of the art results on many classification problems.

In this paper, we propose a two-layer mixture framework for ensemble different networks. In the first layer of this framework, two types of ensemble learning methods, bagging and boosting, are separately applied. Different from traditional methods, our bagging process makes use of the whole training set, not random subsets, to train each model in the ensemble, where several differentiated deep networks are used to promote model variance. To avoid the bias of small-scale samples, the second layer only adopts bagging to mix the results obtained in the first layer. Experimental results demonstrate that the proposed mixture framework performs better than both of the individual network model and independent ensemble method in apparel attributes classification.

## 2    Methodology

### 2.1    Ensemble Framework

The ensemble method is based on the assumption that different classifiers are complementary to each other. There are mainly two strategies to ensemble the predictors: bagging via straightforward weighted voting of the outputs, and boosting through training a new classifier by concatenating feature vectors from the predictors.

The bagging methods are effective for the small number of predictors and require no extra data. However, the ensemble results are highly correlated with the original predictors hence the accuracy cannot be consistently improved with the increase of the number of predictors. Boosting methods, such as XGBoost [10] and CatBoost [1], have been proposed to learn a new model for ensembling weak classifiers. After concatenating the output probabilities of weak classifiers to a vector as the input, a new predictor can be trained through boosting with a set of new training data, which is beneficial to avoiding the risk of overfitting.

Given the fixed amount of data, increasing the training data for ensemble will decrease the training data for CNN predictors. Empirical studies show that when the number of training data for ensemble is small, the performance of bagging and boosting is basically close. It is also observed that boosting produces different results from bagging even with the same predictors, which means that boosting the predictors may result in less correlation with bagging them. Based on this observation we propose a two-layer ensemble framework, as shown in Fig. 1. In the first layer, we pick out some base predictors to compose K groups (possibly overlapped) and use boosting methods to produce K new predictors. The new K predictors have relatively higher accuracy than base predictors after boosting. At the same time, some base predictors with high accuracy are bagged through weighted voting to ensure the robustness of the framework. Different from traditional methods, we trained differentiated deep networks to promote model variance. In this way, the whole training set are used in the bagging process. In the second layer, the bagging method is performed on these new K+1 predictors. In the bagging process, the weight tuning is dependent on the capability of base predictors.

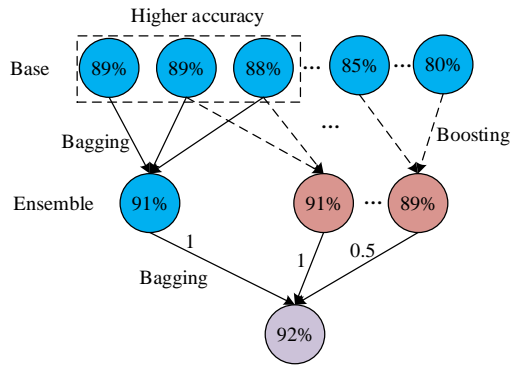

**Fig. 1** Structure of the two-layer mixture network ensemble framework. In the first layer, base predictors are ensembled with bagging and boosting methods respectively. The second layer combines each model with the bagging strategy.

### 2.2 Mixture Network Ensemble

**Resizing Strategy.** For deep networks, input images generally have a fixed size during training. Directly upsampling original images probably changes the aspect ratio of images and deforms the objects in images. As a result, important clues for classification may be lost. For instance, some apparel attributes are with regards to the length of clothes. If the aspect ratio of a long skirt is changed in the process of resizing, the apparel attributes may be predicted as a mid-length skirt. In our resizing strategy, the aspect ratio of original images is fixed during scaling and the scaled images are padded with a certain RGB value for network training.

**Predictor Grouping.** Deep neural networks have shown great potentials in image recognition tasks. In the proposed ensemble framework, diverse deep networks are trained as the base predictors. The networks vary from the network architecture, the optimization method and the resolution of input images. In our implementation, 15 popular network models are utilized as base predictors for ensemble, including three Resnet50 [6] models with 256px, 384px and 512px resolution inputs (1:1 aspect ratio), one Resnet152 model with 512px inputs, one SE-Resnet50 [7] model with 512px inputs, one SE-Resnext50 model with 512px inputs, one SE-Resnext101 model with 384px inputs, two Inception-V4 [8] models with 299px and 512px inputs, two DenseNet121 [9] Models with 256px and 384px inputs and four DenseNet201 Models with 256px,

384px and 512px inputs. All the models are initialized by publicly available ImageNet[5] parameter values.

**Data Augmentation.** To improve the performance of base predictors, we augment original data through random flipping, rotation (-45° ~ +45°), contrast adjustment (0.7 ~ 1.3), random Gaussian blur and so on.

## 3   Experiments

In this section, we evaluate the performance of single model and the ensemble methods on apparel attributes classification. All apparel images from the fashionAI competition are divided into 8 subsets each of which corresponds to an apparel attributes key. In each subset, there are 5000-20000 images with annotated attributes. Apparel attributes classification aims to predict the attributes probability of unknown clothes images. Specifically, the fashionAI competition can be treated as 8 independent and disjoint classification tasks. The overall classification accuracy is computed through averaging the accuracies of these 8 classifiers. According to the protocol of the competition, the accuracy involving Top-1 prediction probabilities is described as the basic precision. In the process of training base predictors, 10% of original images are left out as validation and test data. During the boosting ensemble, the validation data act as the new training data. The 5 fold cross-validation is conducted in our experiments to prevent overfitting as a result of the small dataset.

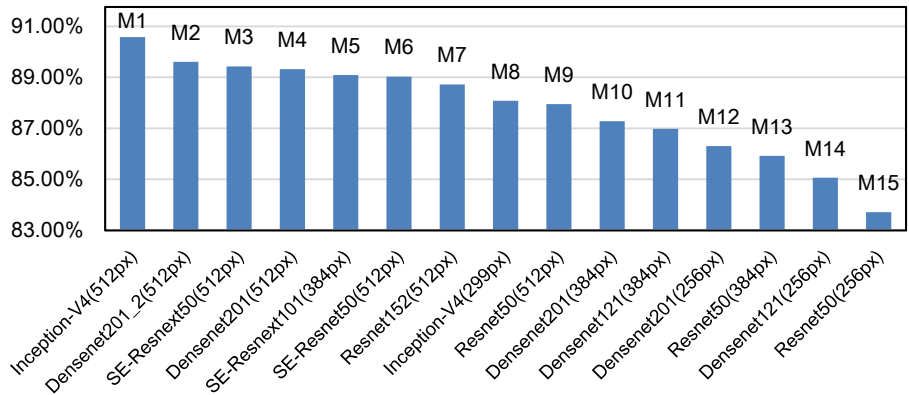

**Fig. 2** Basic precision of 15 base predictors. The predictors are sorted and relabeled in accordance with the accuracy on the validation data.

Fig. 2 illustrates the basic precision of 15 base predictors on the validation data. It is observed that the image resolution networks can accept has a clear effect on the classification accuracy of apparel attributes. In addition, the well-designed network architecture, the reasonable optimization method and the deeper network can improve the accuracy.

For simplicity, our experiments adopt equal weights for base predictors in the bagging ensemble. Two methods, XGBoost and CatBoost, are used in boosting. Fig. 3 illustrates the basic precision with three independent ensemble methods. It is worth noting that the accuracy keeps increasing when the number (N) of predictors is less than 5 (N < 5), but will vibrate with more predictors used. The best accuracy is 92.53%, where predictors M1-M7 are bagged.

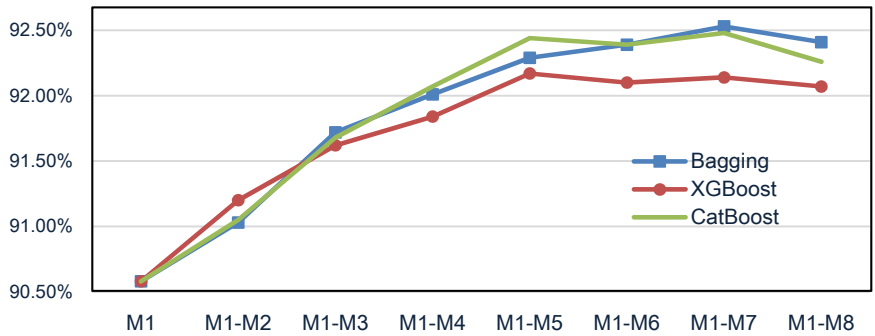

**Fig. 3** Basic precision of three ensemble methods under different number of predictors.

We also computed the difference of predicted labels between the bagging on M1-M7 and other ensemble strategies on M1-M6 or M1-M5. The difference is described as the ratio of the number of different labels to the number of all labels. As presented in Table 1, the boosting produces more differences than the bagging on M1-M6 or M1-M5, where the boosting result is the average of XGBoost and CatBoost. Experiments demonstrate that the results, with other ensemble strategies on M1-M6 or M1-M5, are somewhat different to those with the bagging on M1-M7, even if base predictors are approximate. It is the differentiation that indicates that mixing different strategies can further improve the classification accuracy.

**Table 1** Differences between predicted labels with bagging on M1-M7 and other ensemble strategies on M1-M6 or M1-M5.

|  | M1-M6 | | M1-M5 | |
| --- | --- | --- | --- | --- |
|  | Bagging | Boosting | Bagging | Boosting |
| Difference | 1.23% | **3.17%** | 1.37% | **3.23%** |

As described in Table 2, several mixture strategies in the first layer are tested and compared on the basic precision. If only bagging (i.e., one-layer ensemble) is used, the basic precision is the lowest (92.53%) among them. Other two-layer mixture strategies usually perform better, as shown in the last 3 rows of Table 2. If more base predictors are included through boosting on M8-M12, the basic precision becomes the best (92.76%). This validates the effectiveness of the two-layer mixture network ensemble method in improving the accuracy of apparel attributes classification.

Table 2 Basic precision (BP) under different mixture strategies in the first layer.

| Mixture strategies | BP |
|---|---|
| Bagging | 92.53% |
| Bagging+Boosting(M1-M6) | 92.59% |
| Bagging+Boosting(M1-M6)+Boosting(M1-M5) | 92.69% |
| Bagging+Boosting(M1-M6)+Boosting(M1-M5)+Boosting(M8-M12) | **92.76%** |

## 4      Conclusions

This paper presents a two-layer mixture framework for ensemble different networks. In the first layer of this framework, two types of ensemble learning methods, bagging and boosting, are separately applied. Different from traditional methods using random subsets, the bagging process makes use of the whole training set to train each model in the ensemble, where several differentiated deep networks are used to promote model variance. To avoid the bias of small-scale samples, the second layer only adopts bagging to mix the results obtained with bagging and boosting in the first layer. Experimental results validates the effectiveness of the proposed mixture framework to improve the apparel attributes classification accuracy.